\pdfoutput=1

\documentclass[11pt]{article}

\usepackage{times}
\usepackage{latexsym}

\usepackage[T1]{fontenc}

\usepackage[utf8]{inputenc}

\usepackage{microtype}
\usepackage{ragged2e}
\usepackage{inconsolata}
\usepackage{amssymb}

\title{
Teaching Language Models To Gather Information Proactively
}

\author{
Tenghao Huang\textsuperscript{1\thanks{Work done during Tenghao's internship at Microsoft Office of Applied Research.}} \quad
Sihao Chen\textsuperscript{2} \quad 
Muhao Chen\textsuperscript{3} \quad \\
\textbf{ 
Jonathan May\textsuperscript{1} \quad
Longqi Yang\textsuperscript{2}} 
\textbf{ 
Mengting Wan\textsuperscript{2} \quad
Pei Zhou\textsuperscript{2}}\\ 
\textsuperscript{1}University of Southern California,
\textsuperscript{2}Microsoft Corporation,
\textsuperscript{3}University of California, Davis \\
\texttt{tenghaoh@usc.edu, pei.zhou@microsoft.com}
}

\usepackage[most]{tcolorbox}
\usepackage{inconsolata}
\usepackage{booktabs}
\usepackage{multirow} 
\usepackage{colortbl} 
\usepackage{caption} 
\usepackage[final]{acl}

\usepackage{times}
\usepackage{url}
\usepackage{amsmath} 
\usepackage{cleveref}
\usepackage{latexsym}
\usepackage{graphicx}
\usepackage{array}
\usepackage{tabularx}
\usepackage{enumitem}
\usepackage{adjustbox}
\usepackage{booktabs}
\usepackage{caption}
\usepackage{geometry}
\newcommand{\stitle}[1]{\vspace{1ex} \noindent{\bf #1.}}
\usepackage{xspace}
\usepackage{yfonts}

\usepackage{algorithm}
\usepackage{algorithmic}
\usepackage{microtype}
\crefformat{section}{\S#2#1#3}
\crefformat{subsection}{\S#2#1#3}
\crefformat{subsubsection}{\S#2#1#3}
\crefrangeformat{section}{\S\S#3#1#4 to~#5#2#6}
\crefmultiformat{section}{\S\S#2#1#3}{ and~#2#1#3}{, #2#1#3}{ and~#2#1#3}

\Crefformat{figure}{#2Fig.~#1#3}
\Crefmultiformat{figure}{Figs.~#2#1#3}{ and~#2#1#3}{, #2#1#3}{ and~#2#1#3}
\Crefformat{table}{#2Tab.~#1#3}
\Crefmultiformat{table}{Tabs.~#2#1#3}{ and~#2#1#3}{, #2#1#3}{ and~#2#1#3}
\Crefformat{appendix}{#2Appx.~\S#1#3}
\crefformat{algorithm}{Alg.~#2#1#3}

\definecolor{pb}{HTML}{B0E0E6}
\definecolor{sb}{HTML}{4682B4}       
\definecolor{nv}{HTML}{011180}
\definecolor{or}{HTML}{FF8C00}

\newcommand{\cA}[1]{\cellcolor{pb}#1}
\newcommand{\cB}[1]{\cellcolor{sb}#1}
\newcommand{\cC}[1]{\cellcolor{nv}{\color{white}#1}}   
\newcommand{\cD}[1]{\cellcolor{or}#1}

\begin{document}
\maketitle
\begin{abstract}
Large language models (LLMs) are increasingly expected to function as collaborative partners, engaging in back-and-forth dialogue to solve complex, ambiguous problems. However, current LLMs often falter in real-world settings, defaulting to passive responses or narrow clarifications when faced with incomplete or under-specified prompts—falling short of proactively gathering the missing information that is crucial for high-quality solutions. In this work, we introduce a new task paradigm: proactive information gathering, where LLMs must identify gaps in the provided context and strategically elicit implicit user knowledge through targeted questions.
To systematically study and train this capability, we design a scalable framework that generates partially specified, real-world tasks, masking key information and simulating authentic ambiguity. 
Within this setup, our core innovation is a reinforcement finetuning strategy rewards questions that elicit genuinely new, implicit user information—such as hidden domain expertise or fine-grained requirements—that would otherwise remain unspoken.
Experiments demonstrate that our trained Qwen-2.5-7B model significantly outperforms o3-mini by \textbf{18\%} on automatic evaluation metrics. More importantly, human evaluation reveals that clarification questions and final outlines generated by our model are favored by human annotators by \textbf{42\%} and \textbf{28\%} respectively. Together, these results highlight the value of proactive clarification in elevating LLMs from passive text generators to genuinely collaborative thought partners.

\end{abstract}

\begin{figure}[t]
    \includegraphics[width=\linewidth]{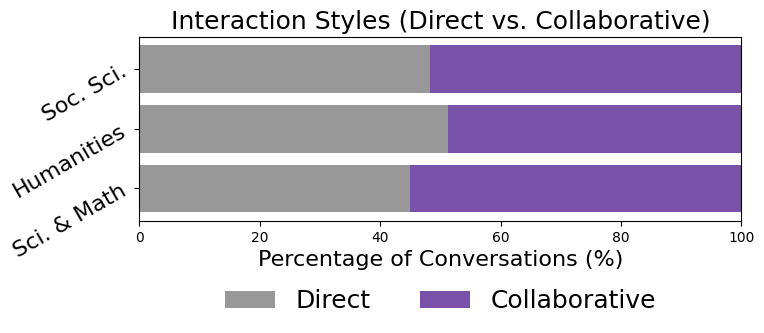}
    
    \caption{Distribution of LLM Interaction Styles Across Disciplines. Collaborative exchanges account for over half of the interactions in all three fields, indicating a widespread shift from one-shot prompting to iterative, multi-turn dialogue \cite{handa2025education}.}
    \vspace{-5mm}
    \label{fig:direct_vs_collab}
\end{figure}

\begin{figure}[h]
    \centering
    \includegraphics[width=0.85\linewidth]{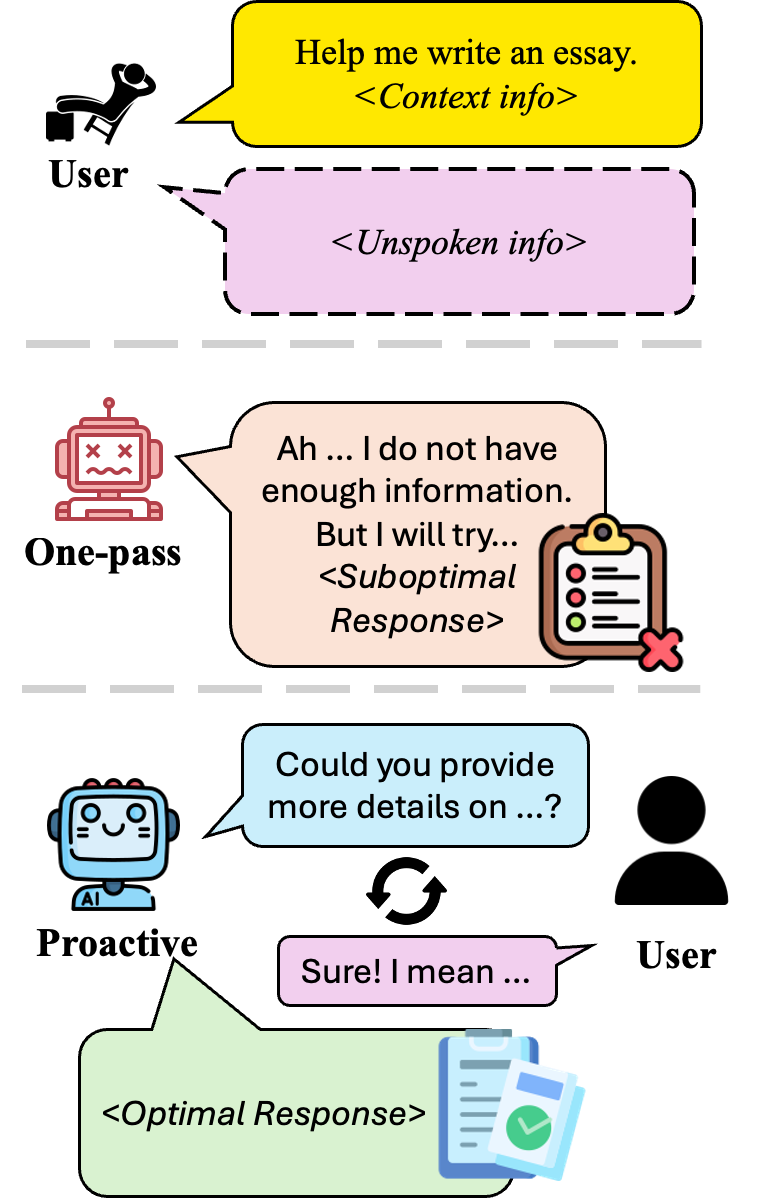}
    
    \caption{Proactive Clarification Enables Optimal LLM Responses.(top) the traditional "one-pass" approach, where the model attempts to respond with limited information, often leading to suboptimal results; (bottom) a proactive approach, where the model detects missing details and engages the user in clarification, ultimately producing a more accurate and helpful response.}
    \vspace{-5mm}
    \label{fig:teaser_fig}
\end{figure}

\section{Introduction}

Large-language models (LLMs) are now indispensable partners for solving diverse reasoning tasks—drafting proofs, debugging code, or writing essays—often excelling when provided with well-structured prompts and sufficient information 
\cite{lightman2023letsverifystepstep, chen2024alphamathzeroprocesssupervision, luo2024improvemathematicalreasoninglanguage, wang2024mathshepherdverifyreinforcellms, 2024largelanguagemonkeys}. Yet, as LLMs become more tightly woven into real-world workflows, the demands on their conversational abilities are evolving. Instead of simply answering queries, users increasingly expect LLMs to act as collaborative partners—engaging in multi-turn, back-and-forth dialogues to tackle complex, ambiguous, and open-ended problems \cite{bommasani2022opportunitiesrisksfoundationmodels, spangher-etal-2025-creative, handa2025education} (\Cref{fig:direct_vs_collab}).

A core challenge in these collaborative settings is information asymmetry: users often provide incomplete or under-specified prompts (“lazy prompting”), and expect the LLM to fill in gaps or steer the conversation productively. As we show in \Cref{tab:qualitative_questions}, off-the-shelf models typically respond with clarifying questions that focus narrowly on what is already present in the current context, or fall back to generic queries that rarely drive the conversation forward.
This approach falls short when the model needs to proactively gather information that is missing, unknown, or unspoken—especially in domains like social science or business, where crucial details are often left unsaid, and no single ``correct'' answer exists.

\textbf{Proactive information gathering} marks a crucial departure from conventional question generation. Rather than merely seeking clarification of ambiguities in existing input, proactive questions poke into missing dimensions, soliciting new, complementary information from the user—details that have not yet surfaced in the conversation. 
The goal is to transform the LLM from a passive responder into a genuine thought partner that can elicit relevant, actionable knowledge, anticipate what is needed, and drive the dialogue toward more productive outcomes (\Cref{fig:teaser_fig}).

However, designing LLMs that excel at proactive information gathering poses several obstacles. High-quality, collaborative dialogue data is scarce and difficult to scale; organic user logs are often noisy and proprietary, while crowd-sourcing nuanced, domain-specific exchanges is expensive and hard to control \cite{malaviya2024dolomitesdomainspecificlongformmethodical, spangher-etal-2025-creative}. Even more fundamentally, the qualities that define a ``good'' proactive question—helpfulness, novelty, contextual complementarity—are subjective and hard to capture with standard reward signals or simple heuristics.

In this work, we address these challenges by designing a framework that specifically rewards pioneering, context-complementary questions—those that reach beyond what is already provided, proactively soliciting critical information from users. Our approach has three core innovations:
\begin{enumerate}[leftmargin=*, nosep]
    \item \textbf{Task Formulation:} We introduce \emph{proactive information gathering} as a new task for LLMs, formalizing the ability to identify and elicit targeted, contextually missing information from users through dialogue.
    \item \textbf{Synthetic Conversation Engine:} Leveraging the \textsc{Dolomites} dataset \cite{malaviya2024dolomitesdomainspecificlongformmethodical}, we construct a simulation pipeline that creates ambiguous prompts and rich clarification trajectories by systematically masking critical information—ensuring that proactive questioning becomes necessary for task completion.
    \item \textbf{Reinforcement Fine-Tuning:} We propose a reward structure that rewards questions reaching beyond the provided context, and fine-tune LLMs using proximal policy optimization. Experiments demonstrate that our trained Qwen-2.5-7B model significantly outperforms o3-mini by \textbf{18\%} on automatic evaluation metrics. More importantly, human evaluation reveals that clarification questions and final outlines generated by our model are favored by human annotators by \textbf{42\%} and \textbf{28\%} respectively. 

\end{enumerate}

Together, these contributions chart a path toward LLMs that are not just compliant responders, but proactive, collaborative partners—able to drive richer, more effective conversations by actively seeking the information that matters most.

\section{Related Work}

\stitle{Proactive Agent}
The term ``\textit{proactive agent}'' has long been correlated with agents that ask follow‑up questions to resolve ambiguities \cite{10.1145/3471158.3472232, 10.1145/3404835.3462839}. However, these efforts focus on addressing slot ambiguities, which aims to clarify ambiguous information present in corpus \cite{guo2021abg, deng2022pacific, pang2024empowering, chen2024learning}. Recent works extend the notion of a `proactive agent', encompassing agents that \textit{predict} user tasks \cite{lu2024proactive}. In this work, our definition of proactive agents entails actively identifying incomplete knowledge regarding domain procedures and user preferences through iterative clarification. 

Recent works emphasize proactive assistance for everyday conversation \cite{chen2024learning}
within fully unobservable environments and reasoning tasks \cite{wu2025collabllmpassiverespondersactive}. In contrast, our work targets open-ended writing tasks and employs a partially observable environment, reflecting real-world scenarios where users inputs imperfect prompt and agents only have access to partial information up front and must strategically clarify hidden details through iterative dialogue.

\stitle{Reinforcement Learning for LLM Alignment}
Previous works focus on reasoning scenarios where step-level rewards are available or the outcome is easy to verify, such as mathematics \cite{chen2024alphamathzeroprocesssupervision, luo2024improvemathematicalreasoninglanguage, lightman2023letsverifystepstep, wang2024mathshepherdverifyreinforcellms}, and coding \cite{2024largelanguagemonkeys} tasks. However, these methods are not generalizable to open-ended tasks, where supervision signals are sparse. Our work proposes a framework that masks user domain knowledge and preferences, providing sufficient learnable reward signals for models at train time.

Another line of works uses reward signals from interactive settings \cite{zhou2023cast, roth2025factored}, typically relying on simulated user feedback to shape the model’s behavior. 

\begin{figure}[t]
    \includegraphics[width=0.85\linewidth]{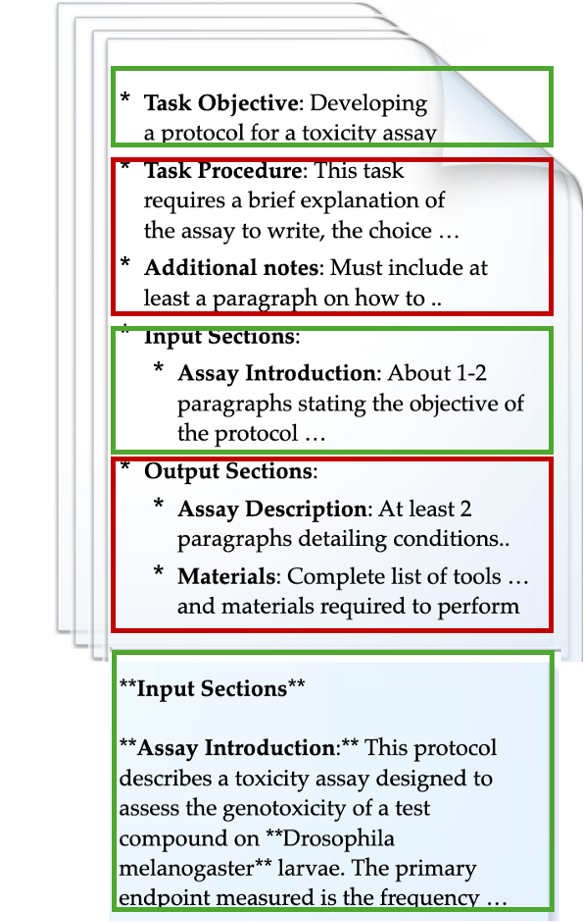}
    
    \caption{An example of our task input. Contents marked by {\textcolor{red}{red}} boxes are not visible to LLMs. Only contents marked by {\textcolor{teal}{green}} boxes are fed to LLMs as input. }
    \vspace{-5mm}
    \label{fig:dolomites_task_example}
\end{figure}

\section{Task and Dataset}
In this section, we formalize the proactive information gathering task (\Cref{ssec: task_definition}), describe how it is instantiated using the \textsc{DOLOMITES} dataset (\Cref{ssec:task_adaption}), and present our evaluation framework (\Cref{ssec: task_evaluation}), which leverages a judge LLM to assess model outputs.

\subsection{Task Definition}
\label{ssec: task_definition}
Let $\mathcal{E}$ denote the \textbf{explicit information} provided by the user (e.g., stated goals, facts, and constraints), and let $\mathcal{I}$ represent the \textbf{implicit information} (unstated assumptions, domain conventions, and fine-grained requirements) necessary for a complete solution. Let $f_\theta$ be a language model parameterized by $\theta$.

The goal is to produce an output $\hat{\mathbf{y}}$ that aligns with the ideal solution $\mathbf{y}^*$, which depends on both explicit and implicit information:
\begin{align*}
  &\text{Ideal\ output:} \quad \mathbf{y}^* = f_\theta(\mathcal{E},\, \mathcal{I}) \\
  &\text{Model\ output:} \quad \hat{\mathbf{y}} =f_\theta(\mathcal{E})
\end{align*}

The objective is to minimize the alignment gap:
\[
\min_\theta\,\, \mathcal{L}\big(\hat{\mathbf{y}},\, \mathbf{y}^*\big)
\]
where loss $\mathcal{L}$ measures deviation from the ideal output.

Since $\mathcal{I}$ is not provided, the assistant must proactively infer or elicit the missing information:
\[
\hat{\mathbf{y}} = f_\theta(\mathcal{E},\, \hat{\mathcal{I}})
\]
where $\hat{\mathcal{I}}$ is the assistant's best estimate of the required implicit information, obtained via clarification or reasoning. 

\subsection{Dataset Adaption}
\label{ssec:task_adaption}
We adapt our data from \textsc{DOLOMITES} datasets \cite{malaviya2024dolomitesdomainspecificlongformmethodical}.
Each instance is a quadruple
\(
\langle \mathbf{o},\mathbf{p},\mathbf{i},\mathbf{s}\rangle
\)
with the following components: 

\begin{enumerate}
  \item Task \textbf{o}bjective ($\mathbf{o}$) — a concise statement of the goal.
  \item Task \textbf{p}rocedure ($\mathbf{p}$) — domain hints or recommended steps.
  \item \textbf{I}nput section ($\mathbf{i}$) — facts, constraints, or numerical data.
  \item Output \textbf{s}pecification ($\mathbf{s}$) — required format, style, and subsections.
\end{enumerate}
The creators elicit 519 task templates from 266 professionals spanning 
\textbf{25 domains} (e.g., medicine, law, civil engineering) and create authentic, real-world writing tasks.

\stitle{Masking scheme}
To align DOLOMITES with our focus on explicit versus implicit information, we adapt each data instance as follows: we first extract the explicit part of each task instance by identifying the task objective ($\mathbf{o}$) and input context ($\mathbf{i}$), which together capture the information a typical user would explicitly provide in a real-world prompt. Formally, 
\[\mathcal{E} = <o,i>.\]
In contrast, we define the implicit part as comprising the procedure or domain expertise ($\mathbf{p}$) and the output specification ($\mathbf{s}$). Formally, 
\[\mathcal{I}=<p,s>.\]
We mask implicit information that is crucial for a high-quality solution but is rarely stated outright by users. During experiments, we simulate incomplete user prompts by revealing only the explicit information ($\mathcal{E}$) to the model, as shown in \Cref{fig:dolomites_task_example}. To successfully complete the task, the model must proactively interact with a user oracle—posing clarification questions to uncover the implicit aspects ($\mathcal{I}$) needed-to produce a satisfactory output.

\stitle{User Response Simulation}
To enable controlled, scalable training and evaluation of proactive information gathering, we implement a user conversation simulation engine that mimics realistic interactions between the assistant and a domain expert. When the model poses a question, the simulated user oracle provides answers based strictly on the masked implicit information—ensuring responses are both faithful to the original task author’s intent and contextually relevant. We will introduce more about this process in \Cref{ssec:synthetic_conversation_engine}.

\subsection{Evaluation Protocol} 

\label{ssec: task_evaluation}
For each instance, we let the output specification be distilled into a set of checklist items $\mathbf{s} = \{c_1, \dots, c_m\}$.
An independent, frozen LLM judge evaluates the assistant’s response $\hat{\mathbf{y}}$ against each checklist item:
\[
\textsc{Match}(c_j, \hat{\mathbf{y}}) \in \{0, 1\},
\]
where $1$ denotes satisfactory coverage. The overall writing score is computed as:
\[
\textsc{Score}(\hat{\mathbf{y}}) = \frac{1}{m} \sum_{j=1}^{m} \textsc{Match}(c_j, \hat{\mathbf{y}}).
\]
No credit is given for omitted or incorrectly formatted items, thereby measuring both content and stylistic fidelity. See \Cref{fig:llm-as-judge} for the judge prompt.

\section{Method}
\label{sec:method}

Our goal is to enable LLMs to \emph{proactively} determine what clarification questions to ask in collaborative problem-solving settings. We frame this as a partially-observable markov decision process (POMDP), and we use reinforcement learning as a way to solve it. 
A central challenge is the absence of reliable, dense supervision for every proposed question. Instead of rewarding questions themselves, we focus on the outcome---\emph{does the question elicit new information that was not already available to the assistant?} This guiding principle motivates our reward formulation, which rewards the discovery of genuinely missing information.

To systematically study and train such proactive clarification behavior, we develop a synthetic conversation engine that simulates realistic, multi-turn assistant–user dialogues (\Cref{ssec:synthetic_conversation_engine}). We then detail our reward design (\Cref{ssec:reward}).

\subsection{Synthetic Conversation Engine}
\label{ssec:synthetic_conversation_engine}

\paragraph{Setting.}
At the start of each episode, the assistant LLM is provided with explicit information
$\mathcal{E}$, while  the implicit information $\mathcal{I}$ is not availble. A second LLM acting as a \textbf{user oracle} has access to both $\mathcal{E}$ and $\mathcal{I}$. Thus, the assistant operates under partial observability.

\paragraph{Dialogue phase.}
Over up to $n$ turns ($n\leq5$ in our experiments), the assistant may ask clarification questions,
\(
q_t \;(t=1,\dots,n),
\)
about the task. The oracle responds with answers
\(a_t = f_\theta(\mathcal{E}, \mathcal{I})\). A good question would help uncover consistent implicit information $(\mathcal{I})$. The running dialogue after $t$ turns is
\(
D_t = \{(q_1, a_1),\dots, (q_t, a_t)\}.
\)

\paragraph{Draft phase.}
$D_t$ can be thought of as $\hat{\mathcal{I}}$, which is the assistant's best estimate of the required implicit information After exhausting the turn budget or issuing a \texttt{STOP}, the assistant must produce its final output,
\(
\hat{\mathbf{y}} = f_\theta(\hat{\mathcal{I}}, D_t).
\)

\subsection{Reward Signal Design}
\label{ssec:reward}

\stitle{Motivation}
Rewarding proactive information gathering is non-trivial because each episode involves two intertwined skills: (\emph{i})~formulating a \emph{useful} question $q$, and (\emph{ii}) evaluating its utility with a reward signal $r$. A naïve approach would be to evaluate the entire final response $\hat{\mathbf{y}}$, incorporating all user answers. However, this is impractical: long-form outputs (typically 500+ tokens) must be scored against a non-exhaustive reference $\mathbf{y}$, resulting in sparse and often uninformative rewards---especially early in training, when useful questions are rare.

\paragraph{Evidence-sentence reward.}
We address this by rewarding the \emph{question} based on whether the answer to the question uncovers genuinely missing information. Intuitively, a valuable clarifying question should:
\begin{enumerate}[nosep,leftmargin=*]
    \item Target information \emph{absent} from the explicit information $\mathcal{E}$,
    \item Be \emph{answerable} using the implicit information $\mathcal{I}$.
\end{enumerate}
To operationalize this, we perform an \emph{evidence-sentence} check at each dialogue turn $t$:

Let the implicit information be split into sentences,
\(\mathcal{I} = \{s_1, \dots, s_{|\mathcal{I}|}\}\),
with indices corresponding to hidden fields
\(\mathcal{H} = \{1, \dots, |\mathcal{I}|\}\). 
Given a question $q_t$, we prompt the user oracle (LLM) to return the set of sentence indices $A(q_t)$ it would cite when answering $q_t$:
\[
A(q_t) = LLM(q_t, \mathcal{I}) \subseteq \{1, \dots, |\mathcal{I}|\}.
\]
The immediate reward is then:
\[
r_t(q_t) =
\begin{cases}
    1, & \text{if } A(q_t) \cap \mathcal{H} \neq \varnothing, \\
    0,  & \text{otherwise}. 
\end{cases}
\]
A question is thus rewarded if it elicits \emph{any} evidence from the hidden fields, incentivizing the model to seek missing information without requiring dense or span-level annotation. We then use this reward signal to train a policy model using Proximal policy optimization (PPO) in an actor–critic framework \cite{schulman2017proximalpolicyoptimizationalgorithms}.






\begin{figure*}[t]
    \includegraphics[width=\linewidth]{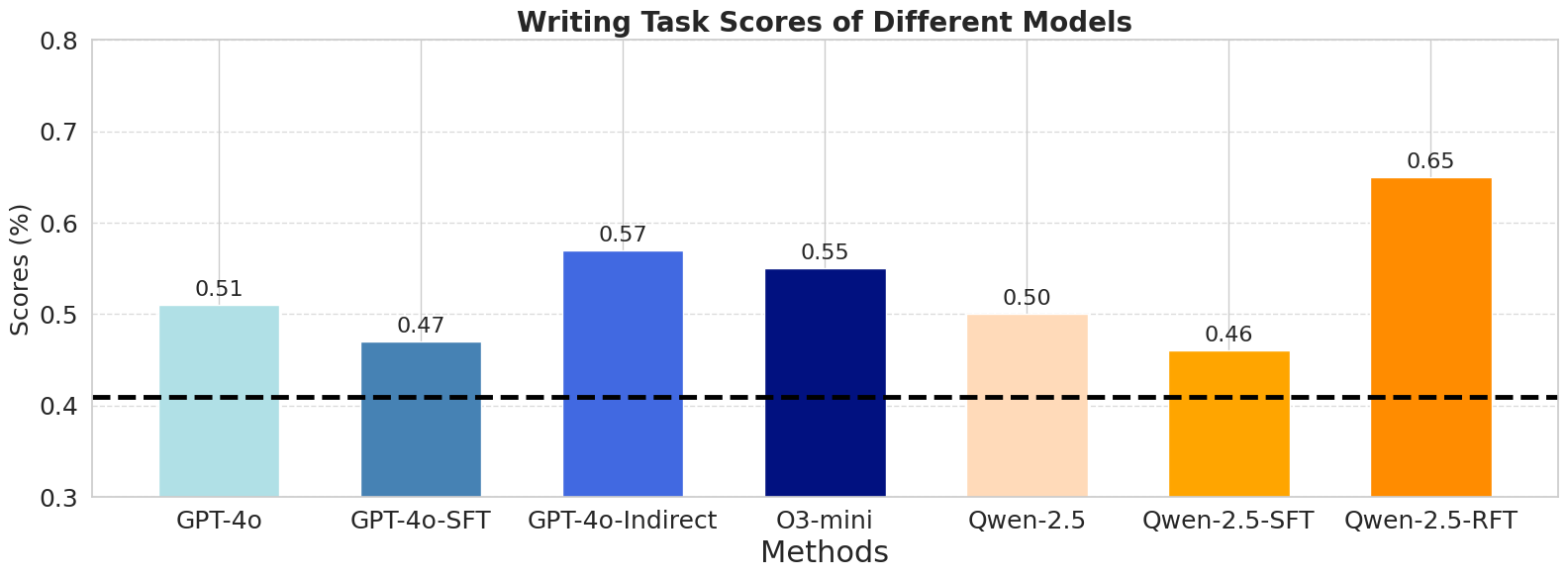}
    
    \caption{Writing Task Performance Across LLM Variants.
Bar chart comparing average writing task scores for several models. Proactively fine-tuned Qwen-2.5-RFT achieves the highest score (0.65), outperforming both base and SFT (supervised fine-tuned) versions of GPT-4o and Qwen-2.5, as well as strong baseline O3-mini. The dashed line marks GPT-4o direct writing performance without asking clarification questions. Our best results (Qwen-2.5-RFT) show statistically
significant difference with baselines, p < 0.05 \cite{koehn2006manual}. }
    \vspace{-2mm}
    \label{fig:main_result}
\end{figure*}

\section{Experiments}
In this section, we evaluate our trained model for proactive information gathering and compare it with baseline methods. We first delve into the details of our experimental setup 
 (\Cref{ssec:impl}), discuss the results obtained (\Cref{ssec:results}), and perform analysis.

\subsection{Implementation Details}
\label{ssec:impl}
We fine-tune a Qwen-2.5-7B model for three epochs, using eight A100 GPUs and the verl implementation of PPO. For each training run, we set a turn budget of five steps per episode. Training is performed with a batch size of 256 episodes, using minibatches of 16. The PPO clipping parameter is set to 0.2. We use separate learning rates for the actor and critic—$2 \times 10^{-5}$ for the actor and $1 \times 10^{-4}$ for the critic. Additionally, we apply gradient norm clipping at 1.0. We fix vanilla GPT-4o as the writer at draft phase throughout our experiments. 

\subsection{Baselines}
\label{sec:baselines}
To isolate the value of \emph{proactive information gathering} we compare our method against four alternative policies, each instantiated with the same GPT-4o or Qwen‑2.5‑7B backbone and evaluated under the dialogue budget $n\!\le\!5$:

\stitle{GPT-4o Direct} The vanilla GPT-4o that \textbf{never} asks clarifying questions. It doesn’t seek to resolve uncertainties or fill in missing details by querying the user. It just generates an output directly.

\stitle{Vanilla LLMs with QA} We employ an in‑context prompting strategy \footnote{Prompt details can be found in \Cref{fig:question_ask_prompt}} that lets the model generate questions to the simulated user in a multi-turn conversation. The conversation will be incorporated before producing the final response. This measures whether a vanilla model's clarification ability can compensate for missing information. Particularly, we include GPT-4o, o3-mini, and Qwen-2.5-7B-Instruct as baseline models.

\stitle{SFT LLMs on Emulated Conversations} Although raw conversations on proactive clarification between users and LLMs are not available, we synthesize multi-turn conversations between models and users, and perform supervised fine-tuning (SFT) with LLMs. Specifically, we prompt LLM with \(
\langle \mathbf{o},\mathbf{p},\mathbf{i},\mathbf{s}\rangle
\) and ask LLM to generate a synthetic conversation if given \(
\langle \mathbf{o},\mathbf{i} \rangle
\), how to uncover \(
\langle \mathbf{p},\mathbf{s}\rangle
\). We use Azure's AI finetuning service\footnote{\url{https://learn.microsoft.com/en-us/azure/ai-foundry/concepts/fine-tuning-overview}} to finetune GPT-4o model and use the \texttt{verl} framework \cite{sheng2024hybridflow} to finetune a Qwen-2.5-7B-Instruct model.

\stitle{GPT-4o Indirectly Supervised} Beyond synthetic data, we realize human brainstorming sessions, though noisy, are also rich in proactive clarification data. QMSUM is a dataset of meeting transcripts in academic, industry and public policy \cite{zhong2021qmsum}. We adapt this dataset, isolating all the conversation turn that proposes questions and train LLM in a DPO fashion \cite{rafailov2024directpreferenceoptimizationlanguage}.

\subsection{Main Results}
\label{ssec:results}

We summarize the performance of all evaluated models on the writing task in \Cref{fig:main_result}. The main findings are as follows:

\stitle{Proactive Clarification Substantially Improves Alignment}
Our method, Qwen-2.5-RFT, achieves the highest score of $0.65$, outperforming all baselines by a large margin. This demonstrates that explicitly training LLMs to proactively ask clarification questions using reinforcement learning leads to significantly better writing task completion, especially in ambiguous or under-specified scenarios.

\stitle{Supervised Fine-Tuning (SFT) Alone is Insufficient}
Both GPT-4o-SFT ($0.47$) and Qwen-2.5-SFT ($0.46$) models, which are fine-tuned on synthetic multi-turn conversations, perform worse comparing to their respective base models (GPT-4o: $0.51$, Qwen-2.5: $0.50$). This indicates that SFT does not robustly endow them with the capacity for contextually proactive questioning in unseen situations, reflecting the challenging nature of our proposed task.

We also observe that GPT-4o-Indirect, trained on real-world brainstorming and meeting data, achieves a competitive score of $0.57$, suggesting that exposure to human-driven clarifications provides useful signal. However, it still underperforms compared to our reinforcement-learned model.

\begin{figure}[t]
    \includegraphics[width=0.95\linewidth]{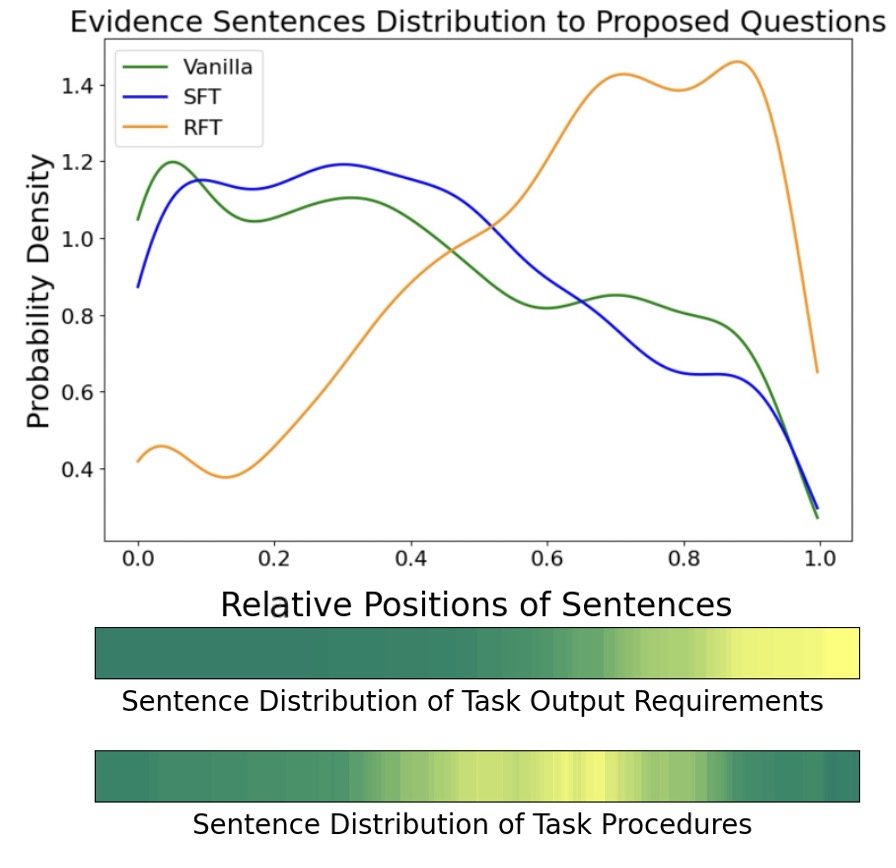}
    
    \caption{Distribution of Evidence Sentences Supporting Model-Generated Questions. The top plot shows the probability density of sentence normalized positions used as evidence for questions proposed by Qwen-2.5-Instruct vanilla, SFT, and RFT models. The two heatmaps indicate where output requirements and task procedures appear within the source documents. The x-axis represents normalized positions within documents (0.0 = beginning, 1.0 = end). Yellow regions in the heatmaps indicate where certain types of information are most densely concentrated.}
    \vspace{-5mm}
    \label{fig:evidence_sentence_distribution_fig}
\end{figure}

\subsection{Further Analysis}
We conduct a comprehensive quantitative analysis to evaluate the effectiveness of proactive clarification training. The results demonstrate clear and consistent gains from our training piepline for proactive question-asking.

\begin{table}[t]
\small
\centering
\setlength{\tabcolsep}{10pt}
\begin{tabular}{@{}lccc@{}}
\toprule
\multirow{2}{*}{\textbf{Method}} & \multicolumn{3}{c}{\textbf{Domains}} \\
\cmidrule(lr){2-4}
& \textbf{Soc. Sci.} & \textbf{Technology} & \textbf{Humanity} \\ \midrule
Direct                 & 0.46 & 0.41 & 0.40 \\ \midrule
\multicolumn{4}{c}{\textit{GPT with 5-turn QA}} \\ \midrule
4o\_Vanilla            & 0.56 & 0.50 & 0.57 \\
4o\_SFT                & 0.62 & 0.36 & 0.45 \\
o3\_mini               & 0.54 & 0.52 & 0.52 \\ \midrule
\multicolumn{4}{c}{\textit{Qwen with 5-turn QA}} \\ \midrule
Vanilla                & 0.65 & 0.48 & 0.45 \\
SFT                    & 0.45 & 0.43 & 0.50\\
RFT                    & \textbf{0.83} & \textbf{0.53} & \textbf{0.71} \\ \bottomrule
\end{tabular}
\caption{Writing task scores of each method across three domains.The largest improvements are seen in social science (+0.37) and humanities (+0.31), underscoring the model’s robustness and ability to proactively clarify under-specified, open-ended tasks that demand deeper contextual reasoning.}
\vspace{-5mm}
\label{tab:scores_divide}
\end{table}

\stitle{Domain-wise Success Rates} \Cref{tab:scores_divide} reports writing task scores across three key domains: social science, technology, and humanities. Our Qwen-2.5-RFT achieves state-of-the-art performance in each domain. 
We also observe that the performance gains are particularly pronounced in domains such as social science and humanities, where the Qwen-2.5-RFT model outperforms the direct baseline by $+0.37$ and $+0.31$ points, respectively. These domains are typically more open-ended and require deeper contextual reasoning, as opposed to technology, which is often more procedural. The strong gains in these complex subjects indicate the strength of our pipeline for tasks demanding nuanced clarification and richer information gathering.

\stitle{Analysis of Evidence Sentence Distributions}
\Cref{fig:evidence_sentence_distribution_fig} examines where in the context models locate their evidence when asking clarifying questions. The Qwen-RFT model demonstrates a clear ability to target both the procedural and output-requirement segments of the context. 
In contrast, the Vanilla and SFT models are more biased toward asking questions on existing information, often failing to pinpoint valuable information in implicit information $\mathcal{I} = \langle \mathbf{p}, \mathbf{s} \rangle$. 
The bottom heatmaps further confirm that the RFT model’s evidence aligns closely with actual distributions of task requirements and procedures, validating its information-seeking behavior. 

\begin{figure}[h]
    \includegraphics[width=\linewidth]{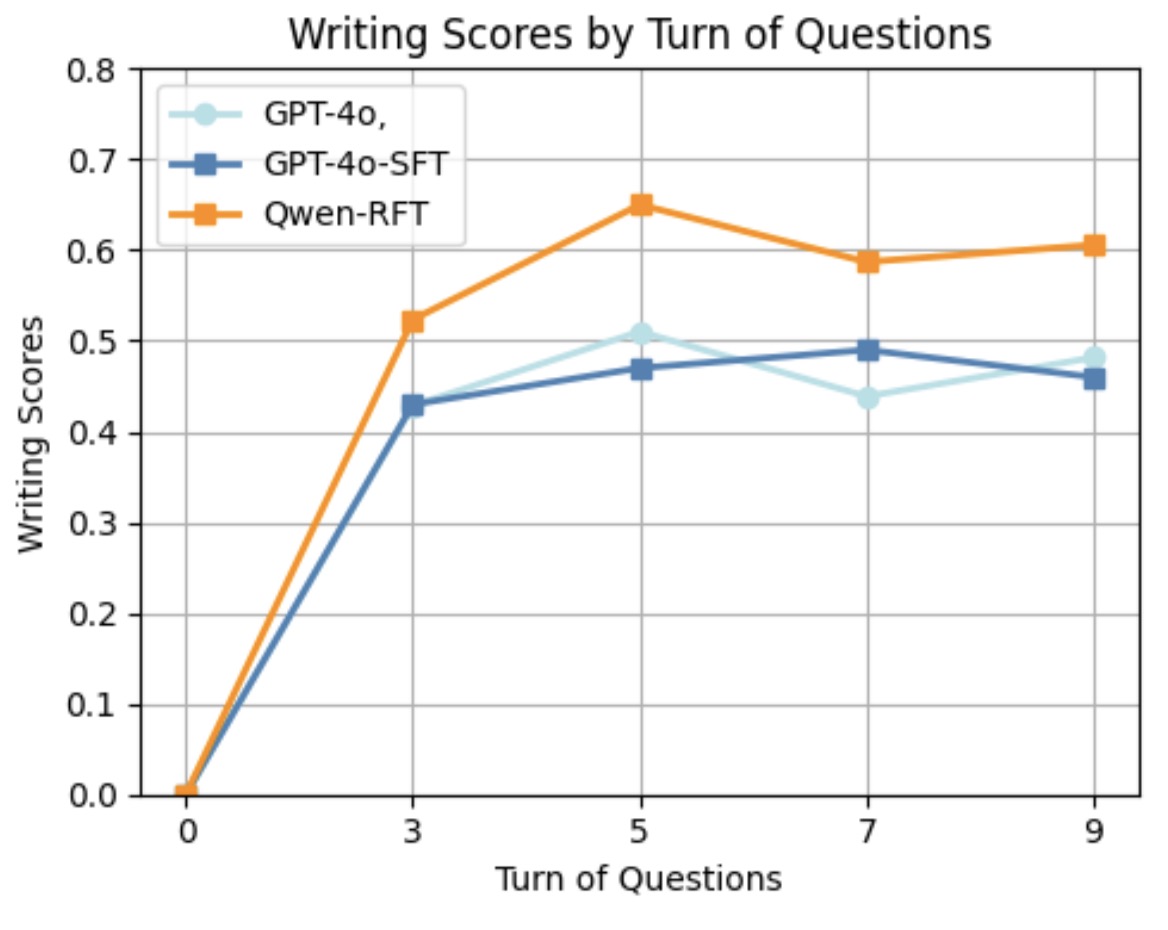}
    
    \caption{Writing task scores across clarification question Rounds. }
    \vspace{-5mm}
    \label{fig:score_by_turn}
\end{figure}


\stitle{Impact of Dialogue Length} \Cref{fig:score_by_turn} reports the impact of question-turn budget on writing performance. While all models benefit from additional clarification rounds, Qwen-RFT exhibits the most pronounced gains, peaking at 5 turns and sustaining high performance with further turns. In contrast, both GPT-4o and GPT-4o-SFT plateau after 3 turns, suggesting diminishing returns without targeted reward optimization.

\begin{table*}[ht]
\small
\centering
\renewcommand{\arraystretch}{1.2}
\setlength{\tabcolsep}{3pt}
\begingroup\sloppy   
\begin{tabularx}{\textwidth}{%
  |>{\RaggedRight\arraybackslash}p{0.30\textwidth}
  |>{\centering\arraybackslash}p{0.15\textwidth}
  |>{\RaggedRight\arraybackslash}X| }
\toprule
\textbf{Context} & \textbf{Model} & \textbf{Clarifying Question} \\ \midrule
\multirow{8}{=}{%
\textbf{task\_objective:} Writing a plan for teaching Maths for Year 2 \par\medskip
\textbf{National curriculum Year 2}\par
$\bullet$ Solve problems with addition and subtraction using concrete objects and pictorial representations.\par
$\bullet$ Apply mental and written methods.\par
$\bullet$ Recall and use facts to 20 fluently and derive related facts to 100.\par\medskip
\textbf{National curriculum Year 1}\par
$\bullet$ Read, write, interpret +, –, =.\par
$\bullet$ Add/subtract one- and two-digit numbers to 20.\par
$\bullet$ Solve one-step problems with concrete objects and pictures.\par\medskip
\textbf{Assessment of previous learning}\par
$\bullet$ Most children add/subtract to 10 confidently.\par
$\bullet$ Some use number bonds; a few still struggle with “equal to”.%
}
& \multirow{2}{*}{\cA{GPT-4o-vanilla}}
  & \cA{How will you assess the students’ progress throughout the plan to ensure they are meeting the learning outcomes?}\\[4pt]
&  \cA{} & \cA{What specific learning outcomes or goals do you want the students to achieve by the end of the Year 2 Maths plan?}\\[4pt]\cline{2-3}

& \multirow{2}{*}{\cB{GPT-4o SFT}}
  & \cB{Do you have any specific learning outcomes or goals you want to achieve with this plan?}\\[4pt]
&  \cB{} & \cB{What activities or resources will you include in Week 2 to advance their learning?}\\[4pt]\cline{2-3}

& \multirow{2}{*}{\cC{o3-mini}}
  & \cC{Could you share which specific maths topics or concepts you’d like to cover for Year 2? For instance, are you focusing on addition and subtraction, basic geometry, or something else?}\\[4pt]
&  \cC{} & \cC{When planning the transition, how would you guide students from concrete objects to pictorial representations—do you have intermediary steps or bridging activities in mind?}\\[4pt]\cline{2-3}

& \multirow{2}{*}{\cD{RFT-Qwen}}
  & \cD{How can I create a flexible lesson-plan framework that adapts to real-time student engagement and understanding during each class session?}\\[4pt]
&  \cD{} & \cD{How can I structure lesson plans to build on existing knowledge while using varied teaching methods for different learning styles and abilities?}\\
\bottomrule
\end{tabularx}
\endgroup

\caption{Clarifying questions generated by each model for the Year 2 maths lesson-planning task.
The prompts probe the three required output sections—\emph{Teaching} (step-by-step teaching plan), \emph{Practice} (pop-up quizzes for in-lesson interactive learning), and \emph{Activities} (reasoning and problem-solving tasks to show mastery for students of different levels).
Notably, \textbf{RFT-Qwen} consistently asks incisive, on-point questions that align closely with these requirements.}

\label{tab:qualitative_questions}
\end{table*}

\stitle{Qualitative Study}
\Cref{tab:qualitative_questions} illustrates sample clarifying questions from different models on the lesson-planning task. We observe that baseline models (GPT-4o-vanilla, SFT, o3-mini) mostly generate generic or surface-level questions, such as asking about learning outcomes or activity details. In contrast, RFT-Qwen consistently produces deeper, more targeted questions—probing for strategies to adapt lessons to student needs and methods to connect prior knowledge to new content. As demonstraed in \Cref{tab:qualitative_questions}, these questions directly support the required output sections (Teaching, Practice, Activities) and demonstrate greater pedagogical insight and context awareness. 

\stitle{Human Evaluation}
We further conduct comprehensive human evaluation. Human evaluation reveals that clarification questions and final outlines generated by our model are favored by human annotators by \textbf{42\%} and \textbf{28\%} respectively. Together, these results highlight the value of proactive clarification in elevating LLMs from passive text generators to genuinely collaborative thought partners. We presents detailed analysis in \Cref{sec: human_eval}.

\begin{table}[t]
\centering
\begin{tabular}{lccc}
\toprule
\rowcolor{white}
\textbf{Model (Questions)} & \textbf{Win} & \textbf{Tie} & \textbf{Lose} \\
\midrule
RFT-Qwen vs o3-mini & 62\% & 18\% & 20\% \\
\bottomrule
\end{tabular}
\caption{Comparison of model-generated clarification questions. A “win” indicates that RFT-Qwen’s question was preferred over o3-mini’s.}
\label{tab:question_comparison}
\end{table}

\begin{table}[t]
\centering
\begin{tabular}{lccc}
\toprule
\rowcolor{white}
\textbf{Model (Outlines)} & \textbf{Win} & \textbf{Tie} & \textbf{Lose} \\
\midrule
RFT-Qwen vs o3-mini & 50\% & 28\% & 22\% \\
\bottomrule
\end{tabular}
\caption{Comparison of model-generated task outlines. A “win” indicates that RFT-Qwen’s outline was preferred over o3-mini’s.}
\vspace{-5mm}
\label{tab:outline_comparison}
\end{table}



\section{Conclusion}
\label{sec:conclusion}
In this work, we address a fundamental shortcoming of current large language models: their inability to proactively seek out missing, contextually relevant information in ambiguous, open-ended tasks. We formalize proactive clarification as a new benchmark challenge, going beyond traditional clarification and slot-filling to embrace a more collaborative and anticipatory role for LLMs. We present a framework for training LLMs to proactively seek missing information, transforming them from passive responders into active thought partners. By leveraging a synthetic conversation engine and reinforcement fine-tuning, our models consistently outperform strong baselines on both automated metrics and human evaluation, especially in open-ended and under-specified domains. Our results highlight the value of targeted reward optimization for collaborative, context-aware dialogue. We hope this work inspires further research into more realistic multi-turn settings and broader domains, paving the way for LLMs that can engage in richer, more productive human–AI collaboration.

\section*{Limitations}
While our framework marks a step forward in proactive clarification for LLMs, several limitations remain. 

\stitle{Focus on single benchmark} Our experiments are conducted solely on the DOLOMITES benchmark. However, we argue this does not undermine the generalizability of our method. DOLOMITES is uniquely comprehensive, spanning 25 professional domains—from humanities and law to technology and medicine—with task instances and evaluation criteria curated by human experts. This diversity ensures our method is evaluated across a wide spectrum of realistic, complex writing scenarios. Moreover, we argue that high-quality benchmarks that combine open-ended writing tasks with fine-grained, expert-driven evaluation criteria remain scarce in the field; DOLOMITES thus provides a strong and meaningful testbed for proactive clarification research.

\stitle{Future research on multi-turn strategy} Our current approach primarily optimizes for single-turn proactive clarification. While this leads to meaningful improvements in both automated and human evaluations, real-world collaboration often involves more complex, multi-turn interactions—including negotiation, iterative refinement, and the management of evolving user goals. We view the extension of our framework to richer, multi-round conversational settings—possibly incorporating strategies for negotiation and dynamic intent alignment—as an important direction for future work.




\bibliography{custom}

\begin{thebibliography}{24}
\providecommand{\natexlab}[1]{#1}

\bibitem[{Bi et~al.(2021)Bi, Ai, and Croft}]{10.1145/3471158.3472232}
Keping Bi, Qingyao Ai, and W.~Bruce Croft. 2021.
\newblock \href {https://doi.org/10.1145/3471158.3472232} {Asking clarifying questions based on negative feedback in conversational search}.
\newblock In \emph{Proceedings of the 2021 ACM SIGIR International Conference on Theory of Information Retrieval}, ICTIR '21, page 157–166, New York, NY, USA. Association for Computing Machinery.

\bibitem[{Bommasani et~al.(2022)Bommasani, Hudson, Adeli, Altman, Arora, von Arx, Bernstein, Bohg, Bosselut, Brunskill, Brynjolfsson, Buch, Card, Castellon, Chatterji, Chen, Creel, Davis, Demszky, Donahue, Doumbouya, Durmus, Ermon, Etchemendy, Ethayarajh, Fei-Fei, Finn, Gale, Gillespie, Goel, Goodman, Grossman, Guha, Hashimoto, Henderson, Hewitt, Ho, Hong, Hsu, Huang, Icard, Jain, Jurafsky, Kalluri, Karamcheti, Keeling, Khani, Khattab, Koh, Krass, Krishna, Kuditipudi, Kumar, Ladhak, Lee, Lee, Leskovec, Levent, Li, Li, Ma, Malik, Manning, Mirchandani, Mitchell, Munyikwa, Nair, Narayan, Narayanan, Newman, Nie, Niebles, Nilforoshan, Nyarko, Ogut, Orr, Papadimitriou, Park, Piech, Portelance, Potts, Raghunathan, Reich, Ren, Rong, Roohani, Ruiz, Ryan, Ré, Sadigh, Sagawa, Santhanam, Shih, Srinivasan, Tamkin, Taori, Thomas, Tramèr, Wang, Wang, Wu, Wu, Wu, Xie, Yasunaga, You, Zaharia, Zhang, Zhang, Zhang, Zhang, Zheng, Zhou, and Liang}]{bommasani2022opportunitiesrisksfoundationmodels}
Rishi Bommasani, Drew~A. Hudson, Ehsan Adeli, Russ Altman, Simran Arora, Sydney von Arx, Michael~S. Bernstein, Jeannette Bohg, Antoine Bosselut, Emma Brunskill, Erik Brynjolfsson, Shyamal Buch, Dallas Card, Rodrigo Castellon, Niladri Chatterji, Annie Chen, Kathleen Creel, Jared~Quincy Davis, Dora Demszky, and 95 others. 2022.
\newblock \href {https://arxiv.org/abs/2108.07258} {On the opportunities and risks of foundation models}.
\newblock \emph{Preprint}, arXiv:2108.07258.

\bibitem[{Brown et~al.(2024)Brown, Juravsky, Ehrlich, Clark, Le, Ré, and Mirhoseini}]{2024largelanguagemonkeys}
Bradley Brown, Jordan Juravsky, Ryan Ehrlich, Ronald Clark, Quoc~V. Le, Christopher Ré, and Azalia Mirhoseini. 2024.
\newblock \href {https://doi.org/10.48550/arXiv.2407.21787} {Large language monkeys: Scaling inference compute with repeated sampling}.

\bibitem[{Chen et~al.(2024{\natexlab{a}})Chen, Liao, Li, and Fan}]{chen2024alphamathzeroprocesssupervision}
Guoxin Chen, Minpeng Liao, Chengxi Li, and Kai Fan. 2024{\natexlab{a}}.
\newblock \href {https://arxiv.org/abs/2405.03553} {Alphamath almost zero: Process supervision without process}.
\newblock \emph{Preprint}, arXiv:2405.03553.

\bibitem[{Chen et~al.(2024{\natexlab{b}})Chen, Sun, Ar{\i}k, and Pfister}]{chen2024learning}
Maximillian Chen, Ruoxi Sun, Sercan~{\"O} Ar{\i}k, and Tomas Pfister. 2024{\natexlab{b}}.
\newblock Learning to clarify: Multi-turn conversations with action-based contrastive self-training.
\newblock \emph{arXiv preprint arXiv:2406.00222}.

\bibitem[{Deng et~al.(2022)Deng, Lei, Zhang, Lam, and Chua}]{deng2022pacific}
Yang Deng, Wenqiang Lei, Wenxuan Zhang, Wai Lam, and Tat-Seng Chua. 2022.
\newblock Pacific: towards proactive conversational question answering over tabular and textual data in finance.
\newblock \emph{arXiv preprint arXiv:2210.08817}.

\bibitem[{Guo et~al.(2021)Guo, Zhang, Reddy, and Alikhani}]{guo2021abg}
Meiqi Guo, Mingda Zhang, Siva Reddy, and Malihe Alikhani. 2021.
\newblock Abg-coqa: Clarifying ambiguity in conversational question answering.
\newblock In \emph{3rd Conference on Automated Knowledge Base Construction}.

\bibitem[{Handa et~al.(2025)Handa, Bent, Tamkin, McCain, Durmus, Stern, Schiraldi, Huang, Ritchie, Syverud, Jagadish, Vo, Bell, and Ganguli}]{handa2025education}
Kunal Handa, Drew Bent, Alex Tamkin, Miles McCain, Esin Durmus, Michael Stern, Mike Schiraldi, Saffron Huang, Stuart Ritchie, Steven Syverud, Kamya Jagadish, Margaret Vo, Matt Bell, and Deep Ganguli. 2025.
\newblock \href {https://www.anthropic.com/news/anthropic-education-report-how-university-students-use-claude} {Anthropic education report: How university students use claude}.

\bibitem[{Koehn and Monz(2006)}]{koehn2006manual}
Philipp Koehn and Christof Monz. 2006.
\newblock Manual and automatic evaluation of machine translation between european languages.
\newblock In \emph{Proceedings of the workshop on statistical machine translation}, pages 102--121. Association for Computational Linguistics.

\bibitem[{Lightman et~al.(2023)Lightman, Kosaraju, Burda, Edwards, Baker, Lee, Leike, Schulman, Sutskever, and Cobbe}]{lightman2023letsverifystepstep}
Hunter Lightman, Vineet Kosaraju, Yura Burda, Harri Edwards, Bowen Baker, Teddy Lee, Jan Leike, John Schulman, Ilya Sutskever, and Karl Cobbe. 2023.
\newblock \href {https://arxiv.org/abs/2305.20050} {Let's verify step by step}.
\newblock \emph{Preprint}, arXiv:2305.20050.

\bibitem[{Lu et~al.(2024)Lu, Yang, Qian, Chen, Luo, Wu, Wang, Cong, Zhang, Lin et~al.}]{lu2024proactive}
Yaxi Lu, Shenzhi Yang, Cheng Qian, Guirong Chen, Qinyu Luo, Yesai Wu, Huadong Wang, Xin Cong, Zhong Zhang, Yankai Lin, and 1 others. 2024.
\newblock Proactive agent: Shifting llm agents from reactive responses to active assistance.
\newblock \emph{arXiv preprint arXiv:2410.12361}.

\bibitem[{Luo et~al.(2024)Luo, Liu, Liu, Phatale, Guo, Lara, Li, Shu, Zhu, Meng, Sun, and Rastogi}]{luo2024improvemathematicalreasoninglanguage}
Liangchen Luo, Yinxiao Liu, Rosanne Liu, Samrat Phatale, Meiqi Guo, Harsh Lara, Yunxuan Li, Lei Shu, Yun Zhu, Lei Meng, Jiao Sun, and Abhinav Rastogi. 2024.
\newblock \href {https://arxiv.org/abs/2406.06592} {Improve mathematical reasoning in language models by automated process supervision}.
\newblock \emph{Preprint}, arXiv:2406.06592.

\bibitem[{Malaviya et~al.(2024)Malaviya, Agrawal, Ganchev, Srinivasan, Huot, Berant, Yatskar, Das, Lapata, and Alberti}]{malaviya2024dolomitesdomainspecificlongformmethodical}
Chaitanya Malaviya, Priyanka Agrawal, Kuzman Ganchev, Pranesh Srinivasan, Fantine Huot, Jonathan Berant, Mark Yatskar, Dipanjan Das, Mirella Lapata, and Chris Alberti. 2024.
\newblock \href {https://arxiv.org/abs/2405.05938} {Dolomites: Domain-specific long-form methodical tasks}.
\newblock \emph{Preprint}, arXiv:2405.05938.

\bibitem[{Pang et~al.(2024)Pang, Fan, Wang, Xiao, Tang, Yang, Jia, Huang, and Yu}]{pang2024empowering}
Jing-Cheng Pang, Heng-Bo Fan, Pengyuan Wang, Jia-Hao Xiao, Nan Tang, Si-Hang Yang, Chengxing Jia, Sheng-Jun Huang, and Yang Yu. 2024.
\newblock Empowering language models with active inquiry for deeper understanding.
\newblock \emph{arXiv preprint arXiv:2402.03719}.

\bibitem[{Rafailov et~al.(2024)Rafailov, Sharma, Mitchell, Ermon, Manning, and Finn}]{rafailov2024directpreferenceoptimizationlanguage}
Rafael Rafailov, Archit Sharma, Eric Mitchell, Stefano Ermon, Christopher~D. Manning, and Chelsea Finn. 2024.
\newblock \href {https://arxiv.org/abs/2305.18290} {Direct preference optimization: Your language model is secretly a reward model}.
\newblock \emph{Preprint}, arXiv:2305.18290.

\bibitem[{Ren et~al.(2021)Ren, Yin, Chen, Wang, Huang, and Zheng}]{10.1145/3404835.3462839}
Xuhui Ren, Hongzhi Yin, Tong Chen, Hao Wang, Zi~Huang, and Kai Zheng. 2021.
\newblock \href {https://doi.org/10.1145/3404835.3462839} {Learning to ask appropriate questions in conversational recommendation}.
\newblock In \emph{Proceedings of the 44th International ACM SIGIR Conference on Research and Development in Information Retrieval}, SIGIR '21, page 808–817, New York, NY, USA. Association for Computing Machinery.

\bibitem[{Roth et~al.(2025)Roth, Hidey, Spangher, Arnold, Ye, Masiewicki, Baek, Grabowski, and Ie}]{roth2025factored}
Nicholas Roth, Christopher Hidey, Lucas Spangher, William~F Arnold, Chang Ye, Nick Masiewicki, Jinoo Baek, Peter Grabowski, and Eugene Ie. 2025.
\newblock Factored agents: Decoupling in-context learning and memorization for robust tool use.
\newblock \emph{arXiv preprint arXiv:2503.22931}.

\bibitem[{Schulman et~al.(2017)Schulman, Wolski, Dhariwal, Radford, and Klimov}]{schulman2017proximalpolicyoptimizationalgorithms}
John Schulman, Filip Wolski, Prafulla Dhariwal, Alec Radford, and Oleg Klimov. 2017.
\newblock \href {https://arxiv.org/abs/1707.06347} {Proximal policy optimization algorithms}.
\newblock \emph{Preprint}, arXiv:1707.06347.

\bibitem[{Sheng et~al.(2024)Sheng, Zhang, Ye, Wu, Zhang, Zhang, Peng, Lin, and Wu}]{sheng2024hybridflow}
Guangming Sheng, Chi Zhang, Zilingfeng Ye, Xibin Wu, Wang Zhang, Ru~Zhang, Yanghua Peng, Haibin Lin, and Chuan Wu. 2024.
\newblock Hybridflow: A flexible and efficient rlhf framework.
\newblock \emph{arXiv preprint arXiv: 2409.19256}.

\bibitem[{Spangher et~al.(2025)Spangher, Huang, Laban, and Peng}]{spangher-etal-2025-creative}
Alexander Spangher, Tenghao Huang, Philippe Laban, and Nanyun Peng. 2025.
\newblock \href {https://aclanthology.org/2025.naacl-tutorial.1/} {Creative planning with language models: Practice, evaluation and applications}.
\newblock In \emph{Proceedings of the 2025 Annual Conference of the Nations of the Americas Chapter of the Association for Computational Linguistics: Human Language Technologies (Volume 5: Tutorial Abstracts)}, pages 1--9, Albuquerque, New Mexico. Association for Computational Linguistics.

\bibitem[{Wang et~al.(2024)Wang, Li, Shao, Xu, Dai, Li, Chen, Wu, and Sui}]{wang2024mathshepherdverifyreinforcellms}
Peiyi Wang, Lei Li, Zhihong Shao, R.~X. Xu, Damai Dai, Yifei Li, Deli Chen, Y.~Wu, and Zhifang Sui. 2024.
\newblock \href {https://arxiv.org/abs/2312.08935} {Math-shepherd: Verify and reinforce llms step-by-step without human annotations}.
\newblock \emph{Preprint}, arXiv:2312.08935.

\bibitem[{Wu et~al.(2025)Wu, Galley, Peng, Cheng, Li, Dou, Cai, Zou, Leskovec, and Gao}]{wu2025collabllmpassiverespondersactive}
Shirley Wu, Michel Galley, Baolin Peng, Hao Cheng, Gavin Li, Yao Dou, Weixin Cai, James Zou, Jure Leskovec, and Jianfeng Gao. 2025.
\newblock \href {https://arxiv.org/abs/2502.00640} {Collabllm: From passive responders to active collaborators}.
\newblock \emph{Preprint}, arXiv:2502.00640.

\bibitem[{Zhong et~al.(2021)Zhong, Yin, Yu, Zaidi, Mutuma, Jha, Awadallah, Celikyilmaz, Liu, Qiu et~al.}]{zhong2021qmsum}
Ming Zhong, Da~Yin, Tao Yu, Ahmad Zaidi, Mutethia Mutuma, Rahul Jha, Ahmed~Hassan Awadallah, Asli Celikyilmaz, Yang Liu, Xipeng Qiu, and 1 others. 2021.
\newblock Qmsum: A new benchmark for query-based multi-domain meeting summarization.
\newblock \emph{arXiv preprint arXiv:2104.05938}.

\bibitem[{Zhou et~al.(2023)Zhou, Zhu, Hu, Pujara, Ren, Callison-Burch, Choi, and Ammanabrolu}]{zhou2023cast}
Pei Zhou, Andrew Zhu, Jennifer Hu, Jay Pujara, Xiang Ren, Chris Callison-Burch, Yejin Choi, and Prithviraj Ammanabrolu. 2023.
\newblock I cast detect thoughts: Learning to converse and guide with intents and theory-of-mind in dungeons and dragons.
\newblock In \emph{Proceedings of the 61st Annual Meeting of the Association for Computational Linguistics (Volume 1: Long Papers)}, pages 11136--11155.

\end{thebibliography}

\appendix
\section{Training Results}
\begin{figure}[h]
    \centering
    \includegraphics[width=\linewidth]{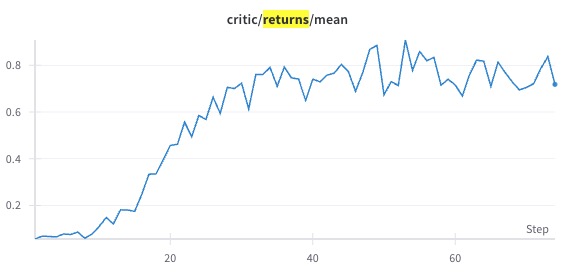}
    
    \caption{Critic return rewards average per step.}
    \vspace{-5mm}
    \label{fig:train_reward}
\end{figure}

\stitle{Validation Reward}
Figure 4 illustrates the learning dynamics of our PPO critic over roughly 70 training steps. The average return remains near zero for the first ten steps—corresponding to a cold‑start phase in which the policy is still exploring—then rises sharply between steps 10 and 35 as the model begins to exploit informative questions and receive consistent positive feedback. After reaching the 0.7–0.8 range, the curve displays a saw‑tooth pattern: returns fluctuate around a high mean, with intermittent peaks above 0.8 that reflect successful episodes.

\begin{figure}[h]
    \includegraphics[width=\linewidth]{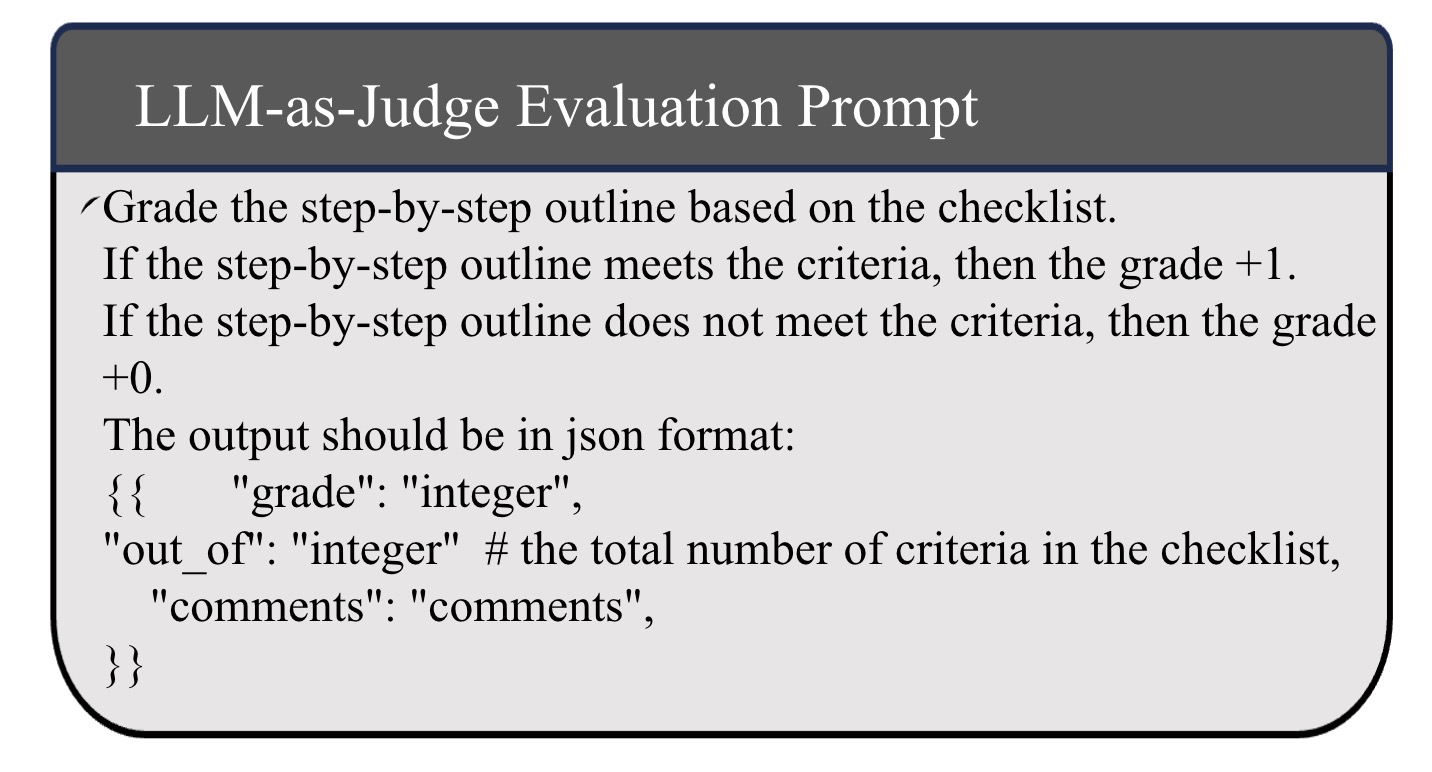}
    
    \caption{LLM-as-Judge style prompt for response evaluation. }
    \vspace{-5mm}
    \label{fig:llm-as-judge}
\end{figure}

\section{Human Evaluation}
\label{sec: human_eval}
To assess the practical utility of the model outputs, we recruited three annotators with graduate-level backgrounds. Annotators were presented with 30 side-by-side outputs from RFT-Qwen and o3-mini for the same input and asked to indicate which model’s output they preferred for question generation and final drafted response. Since full written essay responses are more than 500 words, we instruct the model to generate actionable outlines of response instead that are easier for annotators to evaluate. We perform manual inspection and confirm evaluation of full essay responses and corresponding outlines are equivalent.

We also realize there are questions not directly related to implicit information $\mathcal{I}$ also make sense but not evaluated properly, so we recruit humans to evaluate the question quality by itself. Particularly, for question evaluation, our focus is on brainstorming quality: annotators specifically judged each question for how \emph{helpful} it would be in a brainstorming session. In particular, a good brainstorming question should be \emph{inspiring}—that is, it should encourage creative thinking, open up new perspectives, and invite the user to explore ideas beyond the immediate context. Annotators were instructed to prefer questions that spark further thought, rather than questions that simply clarify surface details. 

For outline evaluation, annotators considered clarity, completeness, and how well the outline meets the task output requirements, which are initially hidden to the models. The results in ~\Cref{tab:question_comparison} and \Cref{tab:outline_comparison} show that RFT-Qwen consistently outperforms o3-mini in both question generation and outline production according to human annotators. Notably, our experiments use \textbf{Qwen-2.5-7B} as the base model for RFT-Qwen, and yet it outperforms o3-mini, a model recognized for its strong reasoning capabilities. This demonstrates that RFT-Qwen’s approach not only produces more human-preferred outputs, but does so even when compared to a larger and well-established reasoning model.

\section{Prompt Details}
\label{sec: prompt_details}
\begin{figure*}[h]
    \includegraphics[width=\linewidth]{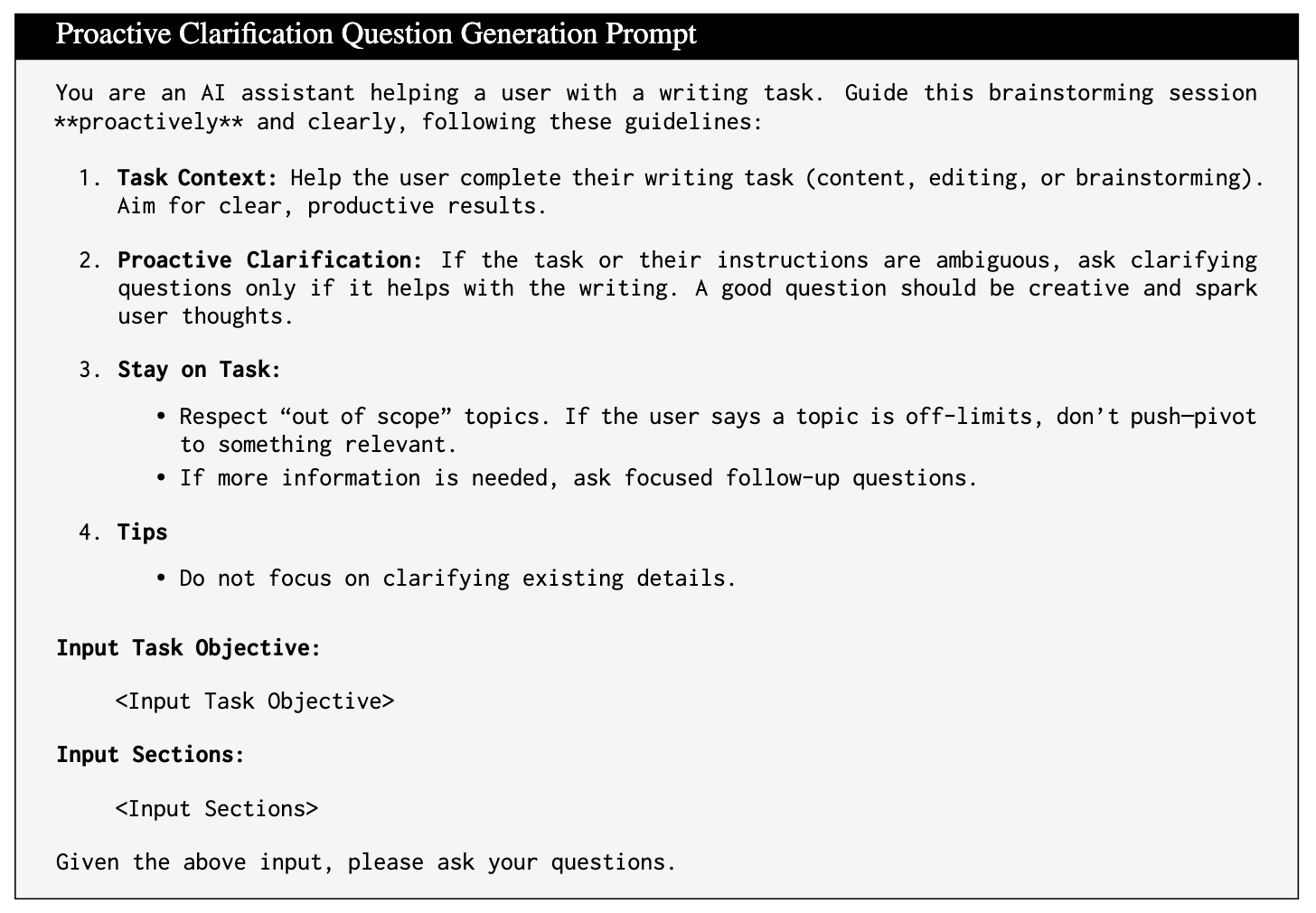}
    
    \caption{Prompt details for proactive clarification question generation.}

    \vspace{-5mm}
    \label{fig:question_ask_prompt}
\end{figure*}

In this section, we showcase prompt details. \Cref{fig:llm-as-judge} shows the LLM-as-judge prompt for evaluation. \Cref{fig:question_ask_prompt} showcases our prompt strategy for instructing models to generate proactive clarification questions.

\end{document}